%% file: main.tex
\documentclass{article} 
\usepackage{iclr2021_conference,times}
\iclrfinalcopy 
\input{math_commands.tex}

\usepackage{hyperref}
\usepackage{url}
\usepackage{multirow}
\usepackage{booktabs}
\usepackage{amsmath,amssymb, mathtools, amsthm}
\usepackage{enumitem}
\usepackage{wrapfig}
\usepackage[normalem]{ulem}
\usepackage{caption}
\usepackage{subcaption}

\newcommand{\Y}{\mathcal{Y}}
\newcommand{\X}{\mathcal{X}}
\newcommand{\f}{f_\theta}

\newcommand{\tableref}[1]{Table~\ref{#1}}

\renewcommand{\eqref}[1]{Eq.~(\ref{#1})}
\newcommand{\figref}[1]{Fig.~(\ref{#1})}

\newcommand{\topmodel}{Best Acc.}

\renewcommand{\[}{\begin{eqnarray}}
\renewcommand{\]}{\end{eqnarray}}
\title{Constant Random Perturbations Provide Adversarial Robustness with Minimal Effect on Accuracy}

\author{Bronya Roni Chernyak \\
Department of Computer Science\\
Bar-Ilan University\\
\texttt{chernroni@gmail.com} \\
\AND
Bhiksha Raj \\
Department of Computer Science\\
Carnegie Mellon University\\
\AND
Tamir Hazan \\
Department of Industrial Engineering and Management\\
Technion \\
\AND
Joseph Keshet\\
Department of Computer Science\\
Bar-Ilan University\\
}
\begin{document}

\maketitle

\begin{abstract}
This paper proposes an attack-independent (non-adversarial training) technique for improving adversarial robustness of neural network models, with minimal loss of standard accuracy. We suggest creating a neighborhood around each training example, such that the label is kept constant for all inputs within that neighborhood. Unlike previous work that follows a similar principle, we apply this idea by {\em extending the training set} with multiple perturbations for each training example, drawn from within the neighborhood. These perturbations are model independent, and remain constant throughout the entire training process. We analyzed our method empirically on MNIST, SVHN, and CIFAR-10, under different attacks and conditions. Results suggest that the proposed approach improves standard accuracy over other defenses while having increased robustness compared to vanilla adversarial training.
\end{abstract}

\input{introduction}
\input{problem_setting}
\input{experiments}

\section*{Acknowledgement}
The work on this paper was partially supported by the Israeli Innovative Authority (WIN5G) and the BIU Center for research in Applied cryptography and Cyber Security in conjunction with the Israel National Cyber Bureau in the Prime Minister’s Office.\\

\bibliography{references.bib}
\bibliographystyle{iclr2021_conference}

\appendix
\input{Appendix}

\end{document}

%% file: math_commands.tex
\usepackage{amsmath,amsfonts,bm}

\def\eqref#1{equation~\ref{#1}}

\def\1{\bm{1}}

\DeclareMathAlphabet{\mathsfit}{\encodingdefault}{\sfdefault}{m}{sl}
\SetMathAlphabet{\mathsfit}{bold}{\encodingdefault}{\sfdefault}{bx}{n}

\DeclareMathOperator*{\argmin}{arg\,min}

%% file: introduction.tex
\section{Introduction}
\label{intro_ewn}
Deep neural networks are susceptible to {\em adversarial attacks} -- the presentation of almost imperceptibly manipulated ``adversarial'' inputs that cause the system to misclassify them \citep{goodfellow2014explaining, moosavi2016deepfool, kurakin2016adversarial,apernot2017practical,fawzi2018analysis, gilmer2019adversarial, papernot2016distillation, cisse2017parseval}.
%
A variety of defense techniques have been proposed in the literature to address this problem. The most successful line of work is {\em adversarial training} \citep{szegedy2013properties, goodfellow2014explaining,  madry2018towards, zhang2019theoretically, carmon2019unlabeled}, which attempts to enhance the robustness of the network by including adversarial examples in the training data itself, with the implicit purpose of ``teaching'' the network not to be fooled by such data instances. Extensions attempt to further improve the robustness through the additional use of pre-processing techniques, on top of the adversarial training \citep{xie2019feature, yang2019me, hendrycks2019using}.

Conventional training views each training example as a \emph{zero volume} element in the space where the output must be correctly predicted. While the continuity of the various activation functions in the network implicitly creates neighborhoods around each training example, such that if a given input falls within this neighborhood, it will be classified correctly \citep{xu2012robustnessvi}, this is not explicitly enforced. Moreover, this neighborhood, might be too small to be robust against adversarial examples. 

Adversarial training explicitly expands the neighborhood around each training example in adversarially sensitive directions. Practically, this is done by selecting an \emph{approximation} of the worst-case adversary: each training example is replaced with an adversarial one that lies within an $\epsilon$ neighborhood of the original input, under some $\ell_p$ norm. Since the adversarially-sensitive directions depend on the model parameters which are continually updated during training, the adversarially modified training instances themselves must be recomputed in each epoch of training. 
This approach achieves increased robustness, often at the cost of standard classification accuracy on non-adversarial data, which becomes more prominent as the input dimension grows \citep{zhang2019theoretically, tsipras2018robustness, carmon2019unlabeled}.

In this paper, we take a different approach and consider each training instance to represent a ball in the input space centered at that instance, such that the label for the instance holds within the entire ball. The ball itself is fixed for each instance, and is not dependent on the model, which is trained to minimize the expected loss over the ball. We approximate this expectation over the ball, by sampling several points within it. This gives us \emph{constant random perturbations} -- a set of perturbations are sampled a priori for every training example from a chosen distribution of an $\epsilon$-bounded domain. We extend the training set with these perturbed examples and keep them constant throughout the training process.

It must be noted that this method differs from conventional adversarial training in two ways: first, unlike adversarial training, the perturbations added to the training set depend only the training instance and do not depend on the model or the adversary. Secondly, also unlike adversarial training where adversarial perturbations must be updated in each iteration of training, the perturbed samples we draw are drawn once and kept fixed during training.

We analyzed the proposed method empirically, under different conditions and under various attacks. We evaluate the standard classification accuracy and adversarial accuracy on MNIST, SVHN, and CIFAR-10. Our method depends on the specific set of perturbed samples added to the training set, and its performance may vary with the specific set of samples drawn. To account for this, for each experiment, we sampled five different perturbation sets and trained a different model. Each result is reported using the average and the standard deviation over the set of models. We also report the result of the best model. We show empirically that using constant perturbation increases robustness compared to standard training while having higher baseline classification accuracy than adversarial training. While practically we could sample only ten perturbations per training example, we show that the robustness could have been further improved using more perturbations. 

Our contribution:
\begin{itemize}[leftmargin=*, topsep=-1pt,noitemsep]
  \item Our approach is attack independent. Compared to previous work, we do not incorporate any adversarial attacks as a part of the training.
  \item Our method is easy to implement. We augment the training set once by adding additional perturbed examples. Other methods  pre-process or manipulate the training data every epoch. Nonetheless, our method is computationally equivalent to the vanilla adversarial training, since the generation of a PGD attack for every example at each iteration is computationally the same as adding one perturbed input to the training set.
  \item Our method has higher classification accuracy than other defenses alongside a higher robustness compared to standard models, and \citet{madry2018towards}. We alleviate robustness by over $50\%$ consistently throughout several datasets: MNIST, SVHN, and CIFAR-10.
\end{itemize}

The implementation of the method is available online\footnote{\url{https://anonymous.4open.science/r/0eb2084e-04ca-488d-92da-05076729a2de/}}.

%% file: problem_setting.tex

\section{Problem Setting}\label{sec:problem_setting}

We denote $\X \subseteq \mathbb{R}^d$ as the set of our input examples, which are represented as $d$-dimensional vectors, and we denote $\Y = \{1,...,M\}$ as the set of class labels. Let $\f: \X \to \Y$ be a neural network with parameters $\theta$, that takes an input from $\X$ and predicts its associated label in $\Y$. During training the parameters $\theta$ are estimated by minimizing the surrogate loss function $\ell(x, y;\theta)$, where $x\in\X$ and $y\in\Y$, over a training set of $M$ examples, $\mathcal{S}=\{(x_i, y_i),~ i = 1\cdots M\}$, as follows:
\begin{equation}
\label{eq:tstar}
\theta^* = \argmin_{\theta}  \sum_{(x,y) \in \mathcal{S}}  \ell(x, y; \theta),
\end{equation}
where $\theta^*$ is the estimated parameters. 


Our goal is to represent each training example by a ball in which the label is kept constant. 
Let $B^{\infty}_{\epsilon}(x)=\{z\in\X : \|z - x \|_\infty \le \epsilon \}$ be the $\ell_\infty$ ball of radius $\epsilon$ around $x$. We define the {\em ball} loss function $\bar{\ell}(x,y;\theta)$ as
\begin{equation}
    \bar{\ell}(x,y;\theta) = \int_{B^{\infty}_\epsilon(0)} \ell(x+\delta,y;\theta) \, d q_\epsilon(\delta)~,
    \label{eq:loss1}
\end{equation}
where $\ell(x,y;\theta)$ is the (point) loss computed at $x$, and $q_\epsilon(\delta)$ is a probability distribution function of the random variable $\delta$. The above Lebesgue integral is generally infeasible to compute; however when $q_\epsilon(\delta)$ is the probability of $\delta$, \eqref{eq:loss1} becomes an expectation of the loss over the ball, {\em i.e.}, $
\bar{\ell}(x,y;\theta) = \mathbb{E}_{\delta\sim q_{\epsilon}} \Big[  \ell(x+\delta,y;\theta) \Big].
$

Since this expectation is infeasible to compute explicitly, we replace it with an empirical average, by drawing $K$ samples from ball around it, according to the distribution $q_\epsilon(\delta)$. Formally, for the training example $(x, y)$, we add a set of perturbed examples $(x+\delta_j, y)$, where $j\in\{1,\ldots,K\}$, $\delta_i \sim q_\epsilon$, and $q_\epsilon$ is a probability distribution function of random variables bounded by $\epsilon$. The network parameters are found by minimizing the average loss as follows:
\begin{equation}
\theta^* = \argmin_\theta \sum_{(x,y) \in \mathcal{S}} \sum_{j=1}^K  \ell(x+\delta_j,y;\theta),
\label{eq:opt1}
\end{equation}
where $\delta \sim q_{\epsilon}$, and $\delta\in[-\epsilon,\epsilon]^d$. 

The samples $(x+\delta_j, y),~j=1\cdots K$ are drawn only once for each training instance, at the outset. This is in contrast to other methods \citep[e.g.,][]{pmlr-v97-cohen19c}, which perturb the inputs at every epoch. We will refer to our approach as \emph{constant perturbations} and the latter sampling techniques as \emph{variable perturbations}. In practice, we aggregate all generated training samples into a single set $\bar{\mathcal{S}}$ of size $KM$, and simply redefine the optimization as
\begin{equation}
\theta^* = \argmin_\theta \sum_{(x,y) \in \bar{\mathcal{S}}}  \ell(x,y;\theta),
\label{eq:opt2}
\end{equation}
Subsequent optimization treats the $KM$ instances in the augmented set as independent training samples, and further training proceeds as usual.

%% file: experiments.tex
\section{Experimental Results}
\label{sec: experiments}

This section presents our empirical evaluation of the proposed method on the CIFAR-10 dataset under different settings (results for MNIST and SVHN are in the Appendix). Our approach is based on constant perturbations that are sampled once at the beginning of the training phase; hence we need to consider the randomness in selecting a good or bad set of perturbations. Therefore, we performed every experiment five times and reported the average accuracy and standard deviation of these experiments, alongside the best model among the five. The perturbations were sampled from a uniform distribution $\mathcal{U}(-16/255,16/255)$. 

We start by presenting our robustness under both white-box and black-box attacks in \tableref{tab:table_cifar10_gen}. We used $\epsilon = 8/255$. The PGD attack was used with a step size  $\eta = 2/255$. We denote by PGD$^n$ the attack created with $n$ iterations. For the black-box setup, we trained 5 models with our method and attacked each one of them with adversarial examples generated from the other 4 models.

\begin{table}[h!]
\vspace{-0.25cm}
\begin{center}
\caption{\it Accuracy of the model trained and tested on CIFAR-10 under different attacks with $\epsilon = 8/255$. \topmodel $~$ represents the accuracy of the best model out of 5.}
\label{tab:table_cifar10_gen}
\footnotesize
\begin{tabular}{lcccc} 

\hline
\hline
\multirow{2}{*}{Setting} & \multicolumn{2}{c}{White-box attack [\%]} & \multicolumn{2}{c}{Black-box attack [\%]} \\
  & Accuracy[\%] & \topmodel [\%]  & Accuracy[\%] & \topmodel [\%] \\
\hline
\hline
No Attack  & $91.56 \pm 0.51$ &  $92.07$ & $91.56 \pm 0.51$ &  $92.07$\\
\hline
FGSM  & $64.09 \pm 2.66$&  $66.75$   & $69.17 \pm 0.58$ &  $69.75$\\
PGD$^{7}$ & $52.22 \pm 4.83$ & $58.93$ & $67.74 \pm 1.63$ &  $69.37$ \\
PGD$^{20}$  & $46.41 \pm 8.00$&  $55.51$ & $66.45 \pm 1.76$&  $68.21$\\
PGD$^{100}$  & $40.76 \pm 11.16$&  $51.92$ & $65.23 \pm 2.20$ &  $67.43$ \\
\hline
\hline
\end{tabular}
\end{center}
\vspace{-0.4cm}

\end{table}

We can observe from the small standard deviation that the standard accuracy of the models is comparable. Similarly, robustness against a weaker adversary such as FGSM, is also similar between models, which is evident from the low variance. Against a stronger adversary such as PGD the gap increases as the number of iterations increases. Nevertheless, if we look at our best model, its robustness is higher than some other models. This indicates the importance of selecting a good sample of the perturbations. 

Additionally, our method's transferability is consistent between the different models, as can be observed from the standard deviation, which is lower for black-box attacks than white-box attack. It is clear that the proposed method adds robustness against the transferability of adversarial examples from other models, in addition to the increased robustness of white-box attacks.

\begin{table}[t!]
\vspace{-0.25cm}
\centering
\caption{\it Comparing results on SVHN of our method with two models of ME-NET
\citep{yang2019me} with different classification accuracy and with \citet{madry2018towards} as reported in \citep{yang2019me}. Attacks are \textbf{white-box} attacks with $\epsilon = 8/255$.}
\label{tab:table_svhn_comp}
\footnotesize
\renewcommand{\arraystretch}{1.2}
\begin{tabular}{lccc} 
\hline
\hline
Method & No attack & PGD$^{20}$ & PGD$^{100}$\\
\hline
Proposed $\mathcal{U}(-8/255,8/255)$\!\!\!  & 94.13 &
74.30 & 74.23\\
Proposed $\mathcal{U}(-16/255,16/255)$ 20 pert.\!\!\!  & \textbf{94.28} &
\textbf{80.32} & \textbf{80.21}\\
Proposed $\mathcal{U}(-16/255,16/255)$\!\!\!  & 93.73 & 69.62 & 69.61\\
\citet{madry2018towards} & 87.40 & 48.40 & 47.50 \\
ME-NET (0.2--0.4) & 87.60 & 75.8 & 69.80\\
ME-NET (0.8--1.0) & 93.50 & 41.40 & 35.50\\
\hline
\hline
\vspace{-0.7cm}
\end{tabular}
\end{table}

We now turn to compare our approach to state-of-the-art methods, which use adversarial training. We start with the SVHN dataset. In \tableref{tab:table_svhn_comp} we state the result of three of our models: our best model trained with $\mathcal{U}(-16/255,16/255)$ with ten and twenty perturbations and with $\mathcal{U}(-8/255,8/255)$\footnote{Additional results on SVHN with different distribution bound are given in Appendix.}. These are compared with \citet{madry2018towards} and ME-NET \citep{yang2019me}\footnote{The results of \citet{madry2018towards} on SVHN are taken from \citet{yang2019me}.}.   We show the performance of two different models of ME-NET, one more robust and one with higher classification accuracy. We can see that our model trained with 20 perturbations has the highest accuracy and robustness. Moreover, in terms of standard accuracy, all of our models have higher accuracy in both settings. In terms of robustness, while for PGD with 20 iterations, our models trained with ten perturbations have lower adversarial accuracy than ME-NET, for PGD with 100 iterations, our model trained with $\mathcal{U}(-8/255,8/255)$ has higher robustness than ME-NET. Furthermore, our model has higher robustness and higher classification accuracy than the ME-NET model with the highest classification accuracy.

\begin{table}[h!]
\begin{center}
\vspace{-0.3cm}
\caption{\it Comparing our method to other methods based on adversarial training on CIFAR-10 using $\epsilon = 8/255$. For PGD we used step size of $2/255$ with 20 iterations Proposed method$^{\text{best}}$ represents the best model out of the five we trained.}
\label{tab:madri_cifar_compcifar}
\footnotesize
\begin{tabular}{lcccc} 
\hline\hline
Method & No Attack & FGSM & PGD$^{20}$ & PGD$^{100}$\\
\hline\hline
Proposed Method$^{\text{best}}$ & \textbf{91.77} & 66.75 & $55.51$ & 51.92\\
 \cite{madry2018towards} & $87.30$ & $56.1$ & $45.80$ & $45.05$\\
 RST$_{adv}$ {\citep{carmon2019unlabeled}} & $89.7 \pm0.1$ & 69.68 & \textbf{63.10} & \textbf{62.14}\\
 TRADES {\citep{zhang2019theoretically}} & $84.92$ & $60.87$ & $55.38$ & 55.13\\
 ME-NET {\citep{yang2019me}} $p: 0.4 \rightarrow 0.6$&  85.50 & \textbf{71.39} & 61.60 & 55.90 \\
  ME-NET {\citep{yang2019me}} $p: 0.6 \rightarrow 0.8$ &  91.00 & - & 58.00 & 53.40 \\
\hline\hline
\end{tabular}
\end{center}
\vspace{-0.4cm}
\end{table}

Next, we turn to compare our performance on CIFAR-10. Results are shown in \tableref{tab:madri_cifar_compcifar}. When comparing our method to other defenses, we should consider that they incorporated additional use of data or computations. Specifically, RST$_{adv}$ \citep{carmon2019unlabeled} used 500k additional unlabeled examples compared to other methods reported. ME-NET \citep{yang2019me} performs matrix estimation on top of adversarial training within the network's pipeline, for every PGD iteration. Additionally, we would like to stress that our method does not incorporate an attack within any step of our training. \begin{wrapfigure}{r}{0.43\textwidth}
\vspace{-0.2cm}
\centering
 \includegraphics[width=0.9\linewidth]{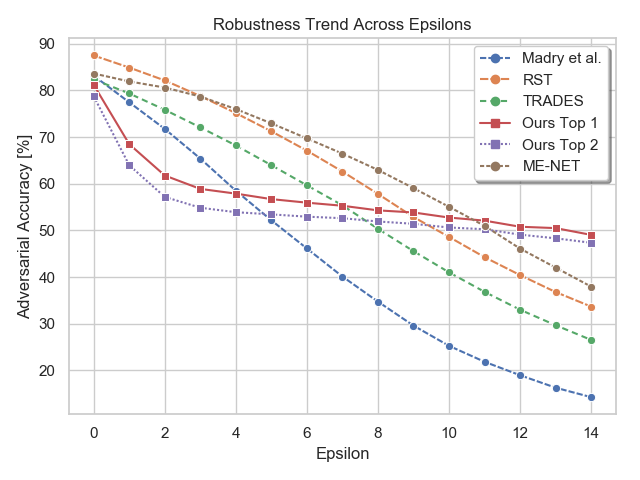}
\vspace{-0.3cm}
\caption{\footnotesize{CIFAR 10: Robustness of our different methods, across different values of $ \epsilon$.}}
\vspace{-0.6cm}
\label{fig:figures}
\end{wrapfigure}We can observe in \tableref{tab:madri_cifar_compcifar} that the classification accuracy (``no attack'') is high, while the robustness of our model is favorably compared to \citet{madry2018towards}, consistently.

Finally, we tested how our model's robustness hold across different values of epsilon against PGD with 20 iterations, and compared it to the robustness of other methods. Results are shown in Fig. 1. It can be observed that for our models, up to $\epsilon = 5$ there is a linear decay in robustness accuracy, and then the decay is more subtle. On the other hand, for models based on adversarial training, the decay remains approximately linear. 


%% file: Appendix.tex
\label{sec: app_experiments}

\section{Supplementary Material}
\subsection{Robustness in a white-box and Black-box setting}
We evaluate our model with different attacks on MNIST, SVHN, and CIFAR-10. We start by describing the architectures used for each of the datasets. These models and hyper-parameters are used in all of the following experiments we present, unless mentioned otherwise.

\textbf{MNIST:} We used LeNet \cite{lecun1995comparison} with two convolutional layers. The first layer has 32 units with 5 filters and the second layer has 64 units with 5 filters. The convolutional layers are followed by ReLU and then by a 2 $\times$ 2 max-pool layer. The two final layers are fully-connected layers, where the first layer is followed by ReLU. We trained the network for 50 epochs and learning rate of 0.001. The perturbations were sampled from a uniform distribution $\mathcal{U}(-0.3,0.3)$. Although our theoretical analysis relies on the perturbation being bounded, we further tested our performance using perturbations sampled from Gaussian distribution  $\mathcal{N}(0,0.3^2)$, as it it has been widely used in previous works.

\textbf{SVHN:} We used ResNet-18 \citep{he2016deep} with adam optimizer, learning rate of 0.002 with a decay, and trained it for 300 epochs.

\textbf{CIFAR-10:} We used Wide-ResNet 28x10 \cite{zagoruyko2016wide}, with dropout of 0.1, learning rate of 0.02 using learning rate decay and trained for 700 epochs. 

Both SVHN and CIFAR-10 perturbations were sampled from a uniform distribution $\mathcal{U}(-16/255,16/255)$. The parameters of the distribution were chosen to be twice the size of the epsilon that was used in the most successful FGSM and PGD attacks against these datasets.

\begin{table*}[h!]
\begin{center}
\caption{\it Accuracy of MNIST with different perturbation distribution and under different attacks with $\epsilon = 0.3$.}
\label{tab:table_mnist_gen}
\begin{tabular}{lcccc} 
\hline
\hline
 & \multicolumn{2}{c}{Adversarial accuracy } & \multicolumn{2}{c}{Adversarial accuracy} \\
Setting & \multicolumn{2}{c}{under white-box attack [\%]} & \multicolumn{2}{c}{under black-box attack [\%]} \\
 \cmidrule(r){2-3} \cmidrule(r){4-5}
& $\mathcal{U}(-0.3,0.3)$ & $\mathcal{N}(0,0.3^2)$ & $\mathcal{U}(-0.3,0.3)$ & $\mathcal{N}(0,0.3^2)$ \\
\hline
No Attack & $99.49 \pm 0.02$ & $99.46 \pm 0.01$ & $99.49 \pm 0.02$ & $99.46 \pm 0.01$ \\
\hline
FGSM & $98.89 \pm 0.05$ & $89.99 \pm 1.35$   & $98.85 \pm 0.02$ & $97.32 \pm 0.49$  \\
PGD$^{100}$ & $99.49 \pm 0.02$ & $95.91 \pm 1.27$ & $98.40 \pm 0.03$ & $98.05 \pm 0.82$\\ 
\hline
\hline
\end{tabular}
\end{center}
\end{table*}

\begin{table}[h!]
\begin{center}
\caption{\it Accuracy of the model trained and tested on SVHN under different attacks with $\epsilon = 8/255$. ``best`` $~$ is the accuracy of the best model (out of 5) with respect to PGD$^{100}$.}
\label{tab:table_svhn_gen}
\begin{tabular}{lccc} 
\hline
\hline
\multirow{2}{*}{Setting} & \multicolumn{2}{c}{White-box} & Black-box \\
 & accuracy[\%] & best [\%] & accuracy[\%] \\
\hline
No attack & $93.64 \pm 0.20$ &  $93.72$\\
\hline
FGSM  & $70.53 \pm 1.32$&  $72.68$& $76.31 \pm 0.094$\\
PGD$^{10}$ & $69.62 \pm 1.45$ & $69.63$ & $74.24 \pm 0.11$\\
PGD$^{20}$ & $67.80 \pm 1.45$&  $69.61$& $73.93 \pm0.08$ \\
PGD$^{100}$  & $67.80 \pm 1.45$&  $69.61$& $73.82 \pm 0.10$ \\
\hline
\hline
\end{tabular}
\end{center}
\end{table}

Results for MNIST, SVHN are presented in \tableref{tab:table_mnist_gen}, and in \tableref{tab:table_svhn_gen} respectively. The classification accuracy of clean images for each dataset is given in the first row of each table. The second column of each table provides results under white-box attacks. For all attacks against MNIST we used $\epsilon = 0.3$ and for all attacks against SVHN we used $\epsilon = 8/255$. For PGD we used a step size of $\eta = 1/255$ for MNIST and $\eta = 2/255$ for SVHN. We denote by PGD$^n$ the attack created with $n$ iterations.

The last column of each table gives the adversarial accuracy under black-box attacks. In the black-box setting, we trained 5 models with our method and attacked each one of them with adversarial examples generated from the other 4 models.

Summarizing these results, we can see that our method attains some resistance to various adversaries. For the MNIST dataset (\tableref{tab:table_mnist_gen}), performance is mostly in favor of the model trained with the uniformly sampled perturbations over the model which was trained with the Gaussian perturbations.
For the SVHN dataset (\tableref{tab:table_svhn_gen}), we can see that the model converges, and in the white-box setting, the adversarial accuracy of PGD with 20 iterations and above remains the same.

Additionally, our method's transferability is consistent between the models, as can be observed from the standard deviation, which is lower for black-box attacks compared to white-box attacks. This consistency is evident for all datasets. It is clear that the proposed method adds robustness against the transferability of adversarial examples from other models, in addition to the increased robustness of white-box attacks.

\subsection{Robustness as a function of the number of perturbations} 
Recall that our objective is to minimize the expected loss within a ball around the training samples. We approximate the expectation through an empirical average over perturbations drawn from the ball. It is to be expected that the performance will depend on the number of perturbations drawn.

In order to see how accuracy and robustness vary with different number of perturbations, we trained models with one, five and ten perturbations. The results are shown in \tableref{tab:table_numpert_mnist}, \tableref{tab:table_svhn_numpert}, and \tableref{tab:table_numpert}. We can observe that using more perturbations increases both accuracy and robustness. Although accuracy is a bit decreased from the standard training model, using more perturbations highly increases robustness by more than $20\%$ for MNIST, $47\%$ for SVHN, and $30\%$ for CIFAR-10, against PGD attack. We can also see that for CIFAR-10 as we increase the number of perturbations, the standard deviation increases between the models, which corresponds to the additional randomness (due to more perturbations) that is being added and is dependent on the sample.

\begin{figure}[h!]
\centering
\hspace{0.3cm}
  \centering
  \includegraphics[width=0.6\linewidth]{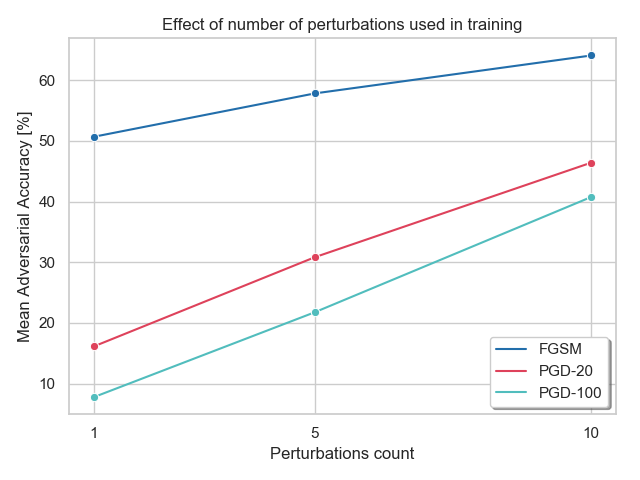}
%
\caption{CIFAR 10: Effect of the number of perturbations on average adversarial accuracy of five models.}
\label{fig:perteffect}

\end{figure}

\begin{figure}[h!]
  \centering
  \includegraphics[width=0.6\linewidth]{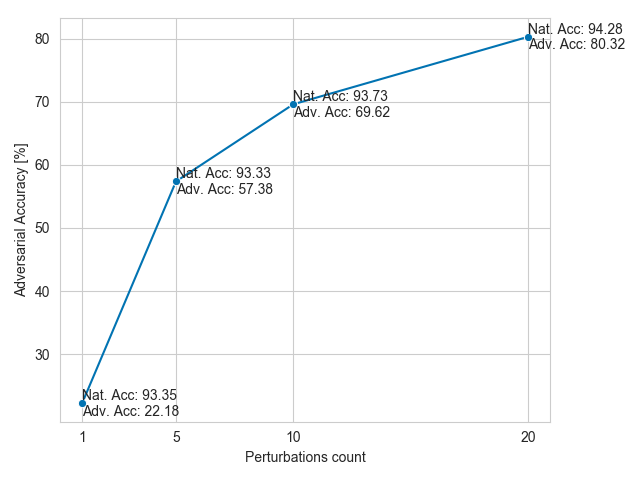}
  \caption{SVHN: Affect of the number of perturbations used in training on adversarial accuracy and standard accuracy. Standard accuracy denoted as ``Nat. Acc''.}
\label{fig:pertcountsvhn}
\end{figure}

Additionally, we plot these results for SVHN in  Fig. \ref{fig:pertcountsvhn} and present
standard accuracy for each perturbation count (denoted ``Nat. Acc'') and plot the results of CIFAR-10 in Fig. \ref{fig:perteffect} respectively. We can observe that for SVHN adversarial accuracy highly increases, by over $50\%$, when using 20 perturbations instead of one, and reached \textbf{80.32\%} accuracy against the attack. Additionally, there is also an increase in the standard accuracy, albeit it is more subtle than the increase in robustness.  We can observe from the plots that \textbf{even using twenty perturbations for SVHN and ten perturbations for CIFAR-10 per example our method did not reach its full potential of the possible robustness.} This trend is consistent through all datasets. This is consistent with the behavior of empirical averages as proxies of expectations, and it may be expected that further increases in the number of samples will improve robustness further. However, naive increase of samples will adversely affect training time. Further research needs to be done to understand how to sample effectively with more perturbations.

\begin{table}[h!]
\begin{center}
\caption{\it The effect of the number of perturbations on robustness for MNIST. The performance of models trained with 1, 5, and 10 perturbations against white-box attacks. PGD attack was created with 100 iterations.}
\label{tab:table_numpert_mnist}
\footnotesize
\begin{tabular}{lcccc} 
\hline
\hline
& \multicolumn{3}{c}{Number of perturbations}\\
Setting & 1 & 5 & 10 \\
\hline
No Attack & $99.43 \!\pm\! 0.05$ & $99.46 \!\pm\! 0.06$ & $99.49 \!\pm\! 0.02$ \\
FGSM & $69.83 \!\pm\! 2.62$ & $97.55 \!\pm\! 0.27$ & $98.89 \!\pm\! 0.05$ \\
PGD$^{100}$ & $50.59 \!\pm\! 1.93$ & $96.24 \!\pm\!  0.39$& $99.49  \!\pm\! 0.02$ \\
\hline\hline
\end{tabular}
\end{center}
\end{table}

\begin{table}[h!]
\begin{center}
\caption{\it The effect of the number of perturbations for SVHN. The performance of models trained with 1, 5, and 10 perturbations against white-box attacks.}
\label{tab:table_svhn_numpert}
\footnotesize
\begin{tabular}{lccc}
\hline\hline
& \multicolumn{3}{c}{Number of perturbations}\\
Setting & 1 & 5 & 10 \\
\hline
No Attack & $93.24 \!\pm\!0.29$ & $93.47 \!\pm\!0.23$ & $93.64  \!\pm\! 0.20$ \\
FGSM & $32.47 \!\pm\! 2.06$ & $62.23 \!\pm\! 0.24$ & $70.53  \!\pm\! 1.32$ \\
PGD$^{10}$  & $21.21 \!\pm\! 2.04$ & $57.14 \!\pm\! 0.37$ &  $69.62 \!\pm\! 1.45$ \\
PGD$^{20}$ &  $20.97 \!\pm\! 2.06$ & $57.11 \!\pm\! 0.36$ & $67.80 \!\pm\! 1.45$ \\
PGD$^{100}$ &  $20.89  \!\pm\! 2.09$ & $57.10  \!\pm\! 0.36$ & $67.80 \!\pm\! 1.45$ \\
\hline\hline
\end{tabular}
\end{center}
\end{table}

\begin{table}[h!]
\begin{center}
\caption{\it The effect of the number of perturbations for CIFAR-10. The performance of models trained with 1, 5, and 10 perturbations against white-box attacks.}
\label{tab:table_numpert}
\footnotesize
\begin{tabular}{lccc}
\hline\hline
& \multicolumn{3}{c}{Number of perturbations}\\
Setting & 1 & 5 & 10 \\
\hline
No Attack & $90.00 \!\pm\! 0.23$ & $90.01 \!\pm\! 1.13$ & $91.56 \!\pm\! 0.51$ \\
FGSM & $50.69 \!\pm\! 3.13$ & $57.84 \!\pm\! 6.50$ & $64.09 \!\pm\! 2.66$ \\
PGD$^{20}$  & $16.13 \!\pm\! 4.96$ & $30.84 \!\pm\! 5.05$ &  $46.42 \!\pm\!8.00$ \\
PGD$^{40}$ &  $11.05 \!\pm\! 3.97$ & $25.56 \!\pm\! 4.87$ & $43.25 \!\pm\! 9.96$ \\
PGD$^{100}$ &  $7.77  \!\pm\! 3.00$ & $21.77  \!\pm\! 4.71$ & $40.76 \!\pm\! 11.16$ \\
\hline\hline
\end{tabular}
\end{center}
\end{table}

\subsection{SVHN: The effect of distribution bound on accuracy and robustness}
\begin{figure}[h!]
  \centering
  \includegraphics[width=0.6\linewidth]{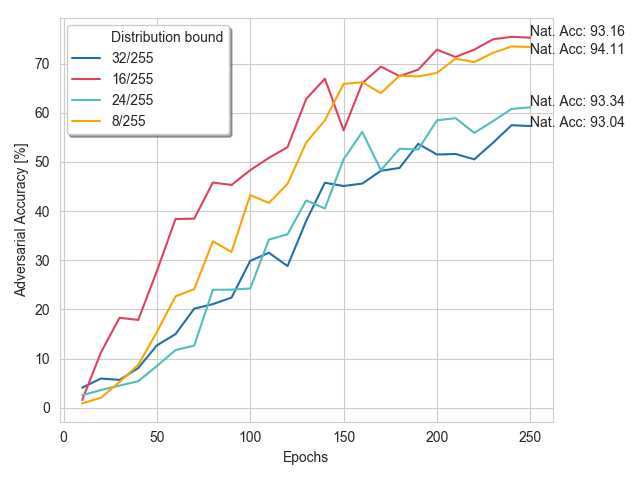}
  \caption{SVHN: the effect of perturbation distributions on accuracy and robustness. Four models were trained with 10 perturbations per image, using different distribution bounds: $\mathcal{U}(-8/255,8/255)$, $\mathcal{U}(-16/255,16/255)$, $\mathcal{U}(-24/255,24/255)$ and $\mathcal{U}(-32/255,32/255)$. The plot shows adversarial accuracy against PGD attack with 20 iterations over the test set. The standard accuracy of the model in the last iteration is denoted as ``Nat. Acc''.}
\label{fig:perteffectsvhn}
\end{figure}

We present the effect of the distribution from which the perturbations were drawn on the accuracy and robustness for SVHN.
In \figref{fig:perteffectsvhn} we show the robustness under PGD attack with 20 iterations and final standard accuracy of models with different distributions: $\mathcal{U}(-8/255,8/255)$, $\mathcal{U}(-16/255,16/255)$, $\mathcal{U}(-24/255,24/255)$ and $\mathcal{U}(-32/255,32/255)$. Standard accuracy of each model is shown at the end of each line plot as ``Nat. Acc''. 

As can be observed, the selection of the perturbation bound is essential. Training with perturbations drawn from $\mathcal{U}(-8/255,8/255)$ achieves the highest standard accuracy and robustness. On the other hand, training with perturbations drawn from $\mathcal{U}(-32/255,32/255)$ deteriorates both standard accuracy and robustness.

\subsection{The effect of constant perturbations compared to variable perturbations}
As mentioned in the introduction, variants on the idea of perturbing inputs to improve robustness are also proposed in \citep{pmlr-v97-cohen19c,lecuyer2019certified}. One of the key distinctions between those approaches and ours is that, in these methods, the noise that is used to perturb the inputs is freshly generated in each iteration. In effect, in contrast to our approach, they employ {\em variable} perturbations that change with iterations.

In order to evaluate the effect of such iteration-dependent perturbations, in this subsection we report several experiments comparing our proposed fixed (drawn once at the start of training) perturbations with iteration-variable perturbations, on CIFAR-10.

We trained classifiers with different perturbation settings: (A) Standard model trained only with the original input images without perturbations; (B) Variable perturbations: model trained only with perturbed images, where the perturbations are sampled anew at each epoch from $\mathcal{U}(-16/255,16/255)$; (C) Model trained with the original input images in addition to their perturbed version as in (B); (D) same as (B) with a normal distribution $\mathcal{N}(0,{16/255}^2)$ (instead of uniform distribution); (E) The original input images along with their perturbed version as in (D); and (G) The original images along with \emph{constant} perturbations drawn from $\mathcal{U}(-16/255,16/255)$ as in our method. G$^{\text{best}}$ indicates G with our best model out of five. The comparison is shown in Table~\ref{tab:table_setup}.

\begin{table*}[t]
\caption{\it Comparison between variable perturbation with different sampling mechanisms and our method. The table presents the performance of models trained with different training methods on CIFAR-10. Models were attacked with $\epsilon=8$ and $\alpha=2$. Training methods:(A) Standart training on clean images only; (B) Training only on perturbed images using Variable perturbations drawn from $\mathcal{U}(-16/255,16/255)$; (C) Training on clean images and their perturbations as in (B); (D) same as (B) where the perturbations are drawn from $\mathcal{N}(0,{16/255}^2)$;(E) Training on clean input images combined with their perturbed version as in (D); (G) Training with our method: constant perturbation drawn from $\mathcal{U}(-16/255,16/255)$. G$^{\text{best}}$ is the best model out of the five shown in G.}
\label{tab:table_setup}
\centering
\renewcommand{\arraystretch}{1.1}
\begin{tabular}{lcccccc} 
\hline\hline
Setting & Accuracy [\%] & FGSM & PGD$^{20}$& PGD$^{40}$ & PGD$^{100}$\\
\hline
A: Standard training & $95.79$ & $46.38$ & $0.30$ & $0.10$ & $0.07 $\\
B: Variable perturb. $\mathcal{U}(-16/255,16/255)$ & $95.17$ & $51.61$ & $8.19$ & $5.68$ & $4.23$ \\
C: (A) \& (B)  & $95.70$ & $62.53$ & $19.71$ & $5.68$ & $4.23$ \\
D: Variable perturb. $\mathcal{N}(0,{16/255}^2)$& \textbf{96.11} & $58.41$ & $2.78$ & $1.56$ & $0.57$ \\
E: (A) \& (D) & $95.00$ & $62.00$ & $25.54$ & $22.39$ & $20.30$ \\
G: Our method & $\!\!91.56 \!\pm\! 0.51\!\!$ & $\!\!64.09 \!\pm\! 2.66\!\!$ &  $\!\!46.41 \!\pm\! 8.00\!\!$ & $\!\!43.25 \!\pm\! 9.96\!\!$ & $\!\!40.76 \!\pm\! 11.16\!\!$ \\
G$^{\text{best}}$: Our best model & $91.77$ & \textbf{66.75} &  \textbf{58.93} & \textbf{55.51} & \textbf{51.92} \\
\hline\hline
\end{tabular}
\end{table*}

%

From \tableref{tab:table_setup} we can observe that while models A-E have higher standard accuracy, in terms of adversarial accuracy against both FGSM and PGD, our method outperforms these models. This is especially true against PGD with more iterations. Albeit, against a weaker adversary such as FGSM, all models increase robustness to some degree. On the other hand, when using more iterations, robustness of most methods with variable perturbations, drops to almost $0\%$.

A possible explanation for the difference in robustness between variable and constant perturbations is that the variance of the cumulative loss function during training is much higher for variable perturbations, where practically the classifier does not see the same example more than once.

Furthermore, we tested the ability of our model to classify perturbed images compared to the model (A), which was trained on the original training set without perturbations. We evaluated the performance of each model on perturbed images from the CIFAR-10 test set. The perturbed test set was created by adding 10 perturbed version for each example, where the perturbations were drawn from $\mathcal{U}(-16/255,16/255)$. The average loss distance of the 10 perturbations was compared to the loss of the original test example. Our model had an average loss difference of 0.036 and variance of 0.035, whereas model (A) had a mean of 0.6 with a large variation of 2.11. Since the perturbations were randomly sampled, this demonstrates that a the loss in the neighbourhood of every test example is flatter when our model is considered.

\begin{table}[ht!]
\begin{center}
\caption{\it {\bf Black-box} attacks against our models on CIFAR-10, using adversarial examples that were {\bf crafted with model A (\tableref{tab:table_setup}) and adversarial training} with $\epsilon = 8/255$.}
\label{tab:table_cifar10_black_baseline}
\renewcommand{\arraystretch}{1.1}
\begin{tabular}{lccc} 
\hline\hline
Source Model:& & Standard & Adv. Training\\
\hline
setting & steps & accuracy [\%] & accuracy [\%]  \\
\hline
FGSM  & - & $79.96 \pm 0.78$ & $76.84 \pm 0.79 $ \\
PGD & 7 & $87.05 \pm 0.82$ & $75.68 \pm 0.79 $\\
PGD & 20 & $87.21 \pm 0.78$& $74.082 \pm 0.87$ \\
PGD & 100 & $88.20 \pm 0.79$& $73.94 \pm 0.81$ \\ 
\hline
\hline
\end{tabular}
\end{center}
\end{table}

\subsection{Black-box attacks from other methods}
We conclude the appendix with our robustness when classifying adversarial examples created from other models.
First, we tested the transferability of adversarial examples from model (A) in \tableref{tab:table_setup}, a baseline model that was trained in standard training only on clean images, to our model. We attacked each of the five models trained with our method using adversarial examples generated with model (A). Additionally, we tested the transferability of adversarial examples generated with a model trained using adversarial training as in \citet{madry2018towards}. Results are shown in \tableref{tab:table_cifar10_black_baseline}.
It is evident that the transferability of adversarial examples from models trained with methods that are different from ours is much lower compared to the one presented in \tableref{tab:table_cifar10_gen}.